\documentclass[final,5p,times,twocolumn]{elsarticle}

\usepackage{lmodern}
\usepackage{amssymb}

\usepackage{amsmath}
\usepackage{epstopdf}

\usepackage{times}
\usepackage{graphicx} 
\usepackage{subfigure} 

\usepackage{natbib}

\usepackage{algorithm}
\usepackage{algorithmic}
\usepackage{color}

\newtheorem{thm}{\rm\bf Theorem}
\newtheorem{lem}[thm]{Lemma}

\newtheorem{coro}[thm]{Corollary}

\newcommand{\dsum}{\displaystyle\sum}
\newcommand{\mbx}{\mathbf{x}}
\newcommand{\mbw}{\mathbf{w}}
\newcommand{\mbz}{\mathbf{z}}
\newcommand{\mby}{\mathbf{y}}
\newcommand{\mbb}{\mathbf{b}}
\newcommand{\mba}{\mathbf{a}}

\newcommand{\dmin}{\displaystyle\min}

\journal{Neural Networks}

\begin{document}

\begin{frontmatter}



\title{ Exploiting Layerwise Convexity of Rectifier Networks with Sign Constrained Weights}


\author{Senjian An$^{1}$, Farid Boussaid$^{2}$, Mohammed Bennamoun$^{1}$ and Ferdous Sohel$^{3}$}

\address{$^{1}$School of Computer Science and Software Engineering\\
       The University of Western Australia\\
       $^{2}$School of Electrical, Electronic and Computer Engineering\\
       The University of Western Australia\\
       $^{3}$School of Engineering and Information Technology\\
 Murdoch University, Australia }

\begin{abstract}

By introducing sign constraints on the weights, this paper proposes sign constrained rectifier networks (SCRNs), whose training can be solved efficiently by the well known majorization-minimization (MM) algorithms. We prove that the proposed two-hidden-layer SCRNs, which exhibit negative weights in the second hidden layer and negative weights in the output layer, are capable of separating any two (or more) disjoint pattern sets. Furthermore, the proposed two-hidden-layer SCRNs can decompose the patterns of each class into several clusters so that each cluster is convexly separable from all the patterns from the other classes. This provides a means to learn the pattern structures and analyse the discriminant factors between different classes of patterns. 

\end{abstract}


\end{frontmatter}



\section{Introduction}\label{sec:introduction}

In recent years, deep rectifier networks have achieved outstanding performance in various applications including object recognition \cite{krizhevsky2012imagenet,zeiler2014visualizing,lee2014deeply,he2015delving}, face verification \cite{taigman2014deepface,sun2014bdeep}, speech recognition (\cite{seide2011conversational,hinton2012deep,deng2013recent} and  handwritten digit recognition \cite{ciresan2012multi}. However, due to their complex hierarchical structures,  deep rectifier networks are not geometrically interpretable and the convergence of their training using stochastic gradient descent methods is still not well understood. Efforts \cite{zeiler2014visualizing} have been made to visualize and understand convolutional layers. Visualization techniques have been used to improve classification performance \cite{zeiler2014visualizing}. However, they can only reveal some general properties of deep neural networks such as the fact that the first layers learn the generic features of images while the last layers learn class-specific features for classification problems.  A recent work  \cite{rister2017piecewiseConvexity} proposes to understand and improve the training of rectifier neural networks using the piecewise convexity property of the objective functions. It proved that, when the objective functions are convex as functions of the outputs of the rectifier neural network, they are piecewise convex as  functions of the parameters of each layer with the other parameters being fixed.   However, there is an exponentially large number of pieces, for which the objective function is convex in each piece but may not be convex across pieces.  
 
 In this paper, we propose sign constrained rectifier networks (SCRNs) and  show that such networks have  some convexity properties, which can be used to develop efficient training algorithms for learning geometrically interpretable classifiers. When the sum of hinge loss and a convex regularisation term is used as the objective function to train the proposed neural networks, the objective function can be minimized using the well known  
majorization-minimization (MM) algorithms \cite{sun2017majorization} (Sec. 5). The MM algorithm is an iterative optimization method  exploiting partial convexities of a function in order to avoid bad local minima and find a good one. The MM algorithm operates by finding a convex surrogate function which upperbounds the objective function. Optimizing the surrogate function drives the objective function downward until a local optimum is reached. For the training of SCRNs, we show that, with any initialization of the parameters, there is a surrogate function which is convex as a function of each layer's parameters when all the other parameters are fixed. Hence, each layer's weights and biases can be learnt alternatively using the MM algorithms.  Furthermore, SCRNs can also decompose each pattern set into several clusters so that each cluster is convexly separable from the patterns of the other classes (Sec. 4). They can thus be used to learn the pattern structures and analyse the discriminant factors between the patterns of different classes. These techniques enable feature analysis for knowledge discovery and for manual supervision to improve the efficiency and performance in training the classifiers. Typical applications include: i) Feature discovery--In health and production management of precision livestock farming \cite{wathes2008precision}, one needs to identify the key features associated with diseases (e.g. hock burn of broiler chickens) on commercial farms, using routinely collected farm management data \cite{hepworth2012broiler}; ii) Supervised shape-free clustering for knowledge discovery--The proposed SCRNs can be used to separate each class of patterns into several clusters (i.e., convex subsets) so that each cluster of the patterns is convexly separable from other classes of patterns, wherein the clusters are not required to be of any particular shape other than convex polytopes; iii) Human-supervised neural network training--The proposed two hidden-layer SCRNs transform the input data into convexly separable data using the first hidden layer. They further transform the data into linearly separable data using the second hidden layer. The decomposition properties of the SCRNs enable human to visualize the patterns, identify the outliers, check the separating boundaries and supervise the training by removing the outliers or mislabelled data. 
\vskip 2mm
{\bf Main Contributions}: In summary, the main contributions of this paper include: 
\begin{itemize}
\item {\bf The introduction of sign constraints on the weights of neural networks in order to learn geometrically interpretable models} (Sec. 2-4). When sign constraints are imposed on the weights of the proposed SCRNs, the first hidden layer transforms the data to be convexly separable while the second hidden layer further transforms the data to be linearly separable. Consequently, every node is a concave (or convex) function of the input of the preceding hidden layer. Since a concave (or convex) piecewise linear function is the minimum (or the maximum respectively) of several linear functions,  the learnt SCRN models are thus geometrically interpretable and can be used to analyse the discriminant features between different classes of patterns.
\item {\bf The introduction of MM algorithms for the training of sign constrained rectifier neural networks} (Sec. 5). The layerwise convexity/concavity properties of the proposed SCRNs result in the existence of a convex surrogate function to upperbound the non-convex hinge loss function so that the efficient MM algorithm can be used to learn the parameters of the neural networks. 
\end{itemize}

{\bf Related Works}: This work is related to \cite{rister2017piecewiseConvexity} which exploits piecewise convexity properties of rectifier neural networks to overcome local minima problems. While \cite{rister2017piecewiseConvexity} uses the piecewise convexity of general rectifier neural networks, this work introduces layer-wise convexity/concavity properties by imposing sign constraints on the weights of the networks, and exploits these properties for pattern decomposition and for efficient training using MM algorithms to reduce the risk of bad local minima.
This work on the universal classification power is related to \cite{hornik1989multilayer,le2010deep,montufar2011refinements}, which address the universal approximation power of deep neural networks for functions  or for probability distributions, and \cite{an2015can} which proves that any multiple pattern sets can be transformed to be linearly separable by two hidden layers, with additional distance preserving properties. In this paper, we prove that any number of pattern sets can be separated by a three-layer (two hidden and one output) neural network with negative weights  in the output layer and negative weights in the second hidden layer. The biases and the weights in the first hidden layer can either be positive or negative. The significance of the proposed SCRNs lies in the fact that it can decompose each class of the patterns into several subsets, where each subset is convexly separable from the other classes of  patterns. This decomposition can be used to analyse pattern sets and identify the discriminant features for pattern recognition. 

Preliminary results of this paper were reported in \citep{an2015sign}, wherein sign constraints were introduced for data decomposition but the discussion was limited to the case of binary classification. This paper extends \citep{an2015sign} to multi-category classification and presents MM-based efficient training algorithms for the proposed SCRNs.

{\bf Notations.} Throughout this paper, we use capital letters to denote matrices, lower case letters for scalar terms, and bold lower letters for vectors. For instance, we use $\mbw_i$ to denote the $i^{th}$ column of a matrix $W$, and use $b_i$ to denote the $i^{th}$ element of a vector $\mathbf{b}$.  For any integer $m$, we use $[m]$ to denote the integer set from 1 to $m$, i.e., $[m]\triangleq \{1,2,\cdots,m\}$. We use $I$ to denote the identity matrix with proper dimensions, $\mathbf{0}$ to denote a vector with all elements being 0, and $\mathbf{1}$ to denote a vector with all elements being 1.  $W\succeq 0$ and $\mathbf{b}\succeq 0$ denote that all elements of $W$ and $\mathbf{b}$ are non-negative while $W\preceq 0$ and $\mathbf{b}\preceq 0$ denote that all elements of $W$ and $\mathbf{b}$ are non-positive. Given a finite number of points $\mbx_i$ ($i\in [m]$) in $\mathbb{R}^n$, a convex combination $\mbx$ of these points is a linear combination of these points, in which all coefficients are non-negative and sum to 1. The convex hull of a set $\mathcal{X}$, denoted by $\mathrm{CH}(\mathcal{X})$, is a set of all convex combinations of the points in $\mathcal{X}$.

{\bf Organization.} The rest of this paper is organised as follows. We introduce  rectifier neural networks with sign constrained weights in Section 2, and investigate the capacity of sign constrained single hidden layer rectifier neural networks for classification and pattern decomposition in Section 3. Section 4 investigates the universal classification power and pattern decomposition capacity of two hidden layer rectifier neural networks with sign constraints on the output layer and on the last hidden layer. Such sign constraints can be used to control the strategy how a two-hidden-layer neural network to achieve linear  separability. In Section 5, we first introduce the general MM algorithm, and then presents how MM algorithms can be used to train sign constrained neural networks.  Section 6 concludes this paper.

\section{Rectifier Neural Networks with Sign Constrained Weights}

A hidden layer in a rectifier neural network can be described by a rectified linear unit (ReLU) as below
\begin{equation}
\mathrm{ReLU}(\mbx;W,\mathbf{b})\triangleq \max(\mathbf{0}, W^T\mbx+\mathbf{b})
\end{equation}
where $W$ is the weight matrix and $\mathbf{b}$ is the bias vector. $\mathrm{ReLU}(\mbx;W,\mbb)$ is a simple yet powerful nonlinear transformation wherein the nonlinearity is imposed by the simplest nonlinear function $\max(0,x)$.  

A rectifier neural network with $m$ hidden layers  can be described as a chain of $m$ ReLUs.
\begin{equation}
\begin{array}{ll}
({\bf \mathrm{ReNN}})& 
\left\{\begin{array}{rcl}
\mbz_1&\triangleq& \mathrm{ReLU}(\mbx;W_1,\mathbf{b}_1)\\
\mbz_k&\triangleq& \mathrm{ReLU}(\mbz_{k-1};W_k,\mathbf{b}_k), 2\leq k\leq m\\
\mby & \triangleq & A^T \mbz_m+\mathbf{c}
\end{array}\right.
\end{array}
\label{ReNN}
\end{equation}
where $\mbx$, $\mby$ and $ \mbz_k$ are the input, the final output and the output of the $k^{th}$ hidden layer respectively. 

 In particular, this paper considers a special class of rectifier neural networks with sign constraints on the weights of the output layer and the second hidden layer. The sign constraints are used to decompose the pattern sets for discriminate factor analysis. 

A single hidden layer sign constrained ReNN imposes non-positiveness on the weights in the output layer and is defined as below: 
\begin{equation}
\begin{array}{ll}
({\bf \mathrm{SCReNN1}})& 
\left\{\begin{array}{rcl}
\mbz&\triangleq& \mathrm{ReLU}(\mbx;W,\mathbf{b})\\
\mby & \triangleq & A^T \mbz+\mathbf{c}\\
A&\preceq& 0
\end{array}\right.
\end{array}
\label{SCReNN1}
\end{equation}
where $\mbx\in\mathbb{R}^n, \mby\in\mathbb{R}^m$ are the input and the output respectively, $W\in\mathbb{R}^{n\times l},A\in\mathbb{R}^{l\times m}$ are the weight matrices  and $\mathbf{b}\in\mathbb{R}^{l}, \mathbf{c}\in\mathbb{R}^m$ are the bias vectors in the hidden layer and output layer respectively.

{
{\it Remark}: In \citep{an2015sign}, non-negativeness on the output layer weights was imposed for sign-constrained rectifier neural networks. In this paper, we use non-positive constraints for convenience in presenting the decomposition properties of such sign constrained ReNN for multiple category classification problems.  
}

For two hidden layer sign constrained ReNNs, we impose non-negativeness on the weights of the output layer and impose non-positiveness on the weights of the second hidden layer. A two hidden layer sign constrained ReNN can be described as below.
\begin{equation}
\begin{array}{ll}
({\bf \mathrm{SCReNN2}})& 
\left\{\begin{array}{rcl}
\mbz_1&\triangleq& \mathrm{ReLU}(\mbx;W_1,\mathbf{b}_1)\\
\mbz_2&\triangleq& \mathrm{ReLU}(\mbz_1;W_2,\mathbf{b}_2)\\
\mby & \triangleq & A^T \mbz_2+\mathbf{c}\\
W_2&\preceq& 0 \\
A&\preceq& 0
\end{array}\right.
\end{array}
\label{SCReNN2}
\end{equation}
where $\mbx\in\mathbb{R}^n, \mby\in\mathbb{R}^n$ are the input and output , $W_1\in\mathbb{R}^{n\times l_1},W_2\in\mathbb{R}^{l_1\times l_2}, A\in\mathbb{R}^{l_2\times m}$ are the weight matrices  and $\mathbf{b}_k\in\mathbb{R}^{l_k} (k=1,2), \mathbf{c}\in\mathbb{R}^m$ are the bias vectors in the first hidden layer, the second hidden layer and the output layer.

Next, we present the properties of sign constrained rectifier networks. A real valued function $f(\mbx)$ from 
$\mathbb{R}^n$ to $\mathbb{R}$ is called a {\it convex} function if 
\begin{equation}
f(\lambda \mbx_1+(1-\lambda)\mbx_0)\leq \lambda f(\mbx_1)+(1-\lambda) f(\mbx_0)
\end{equation}
holds for any $\lambda\in [0,1],\mbx_1,\mbx_0\in\mathbb{R}^n$. $f(\mbx)$ is called concave if $-f$ is convex. The Lemma below addresses the relationship between the convexity (or concavity) of a single hidden layer rectifier network and the signs of its weights in the output layer.

 \begin{lem} {\it Let $f(\mbx;\mba,W,\mbb,c)=\mathbf{a}^T\max\{0,W^T\mbx+\mathbf{b}\}+c$ be a real-valued function from $\mathbb{R}^n$ to $\mathbb{R}$. Then the following statements are true:
 \begin{enumerate}
 \item[(i)] When the parameters $\mba,W,\mbb,c$ are fixed, $f(\mbx;\mba,W,\mbb,c)$ is a convex (or concave) function of $\mbx$ if $\mathbf{a}\succeq 0$ (or $\mathbf{a}\preceq 0$ respectively).\\
 \item[(ii)] When $\mbx,\mba$ are fixed, $f(\mbx;\mba,W,\mbb, c)$ is a convex (or concave) function of $W$ and $\mbb$ if $\mathbf{a}\succeq 0$ (or $\mathbf{a}\preceq 0$ respectively).\\
 \end{enumerate}  
 }
 \label{convexSHL}
 \end{lem}
 
{\bf Proof}: We only need to prove the first statement $(i)$. The proof of $(ii)$ is similar and thus omitted. Denote $\mathbf{a}=[a_1,a_2,\cdots,a_m]$, $W=[\mathbf{w}_1,\mathbf{w}_2,\cdots,\mathbf{w}_m]$ and $\mathbf{b}=[b_1,b_2,\cdots,b_m]$, then 
\begin{equation}
f(\mbx)=\sum_{i=1}^m a_i\max\{0,\mathbf{w}_i^T\mbx+b_i\}+c.
\end{equation}
Let $\mbx_0$ and $\mbx_1$ be two points in $\Omega$, i.e., $f(\mbx_0)>0$ and $f(\mbx_1)>0$, and let 
$\mbx_{\lambda}=\lambda \mbx_1+(1-\lambda)\mbx_0$. Denote $z_i(\lambda)=\mbw_i^T\mbx_{\lambda}+b_i$. Note that 
\begin{equation}
\max\{0,z_i(\lambda)\}\leq \lambda \max\{0,z_i(0)\}+(1-\lambda)\max\{0,z_i(1)\}
\end{equation}
holds for any $\lambda\in [0,1]$. When $a_i\geq 0$ for any $i\in [m]$, we have 
\begin{equation}
\begin{array}{rcl}
f(\mbx_{\lambda})&=& \displaystyle\sum_{i=1}^m a_i\max\{0,z_i(\lambda)\}+c\\
&\leq& \lambda\displaystyle\sum_{i=1}^m a_i\max\{0,z_i(1)\}\\
&&+(1-\lambda)\displaystyle\sum_{i=1}^m a_i\max\{0,z_i(0)\} +c\\
&=& \lambda f(\mbx_1)+(1-\lambda)f(\mbx_0)\\
\end{array}
\end{equation}
for any $\lambda\in [0,1]$, and this implies that $f(\mbx)$ is convex when $\mathbf{a}\succeq 0$. 
\begin{flushright}
$\Box$
\end{flushright}
From Lemma \ref{convexSHL}, we have the following two corollaries for the properties of sign constrained rectifier networks. 
\begin{coro}
{\it Let $\mathrm{SCReNN1}$ be defined as in Eq. (\ref{SCReNN1}). Then every element of the output $\mathbf{y}$ is a concave function of $\mbx$. 
}
\end{coro}

\begin{coro}
{\it Let $\mathrm{SCReNN2}$ be defined as in Eq. (\ref{SCReNN2}). Then every element of the output  $\mathbf{y}$ is a concave function of $\mbz_1$ (i.e. the output of the first hidden layer).
}
\end{coro}

There are two useful properties of convex/concave functions. First, when a classifier is a convex/concave function, it separates the domain into two regions with one being a convex set:  $\{\mbx: f(\mbx)<0\}$ is convex when $f(\mbx)$ is a convex function, and $\{\mbx: f(\mbx)>0\}$ is convex when $f(\mbx)$ is concave. Second, a convex/concave function can be approximated by a series of linear classifiers. These two properties make the sign-constrained rectifier networks more geometrically interpretable. In Section 3 and Section 4, we will investigate how to use these properties to decompose the data for discriminant factor analysis.

\section{The Capacity of  Sign Constrained Rectifier Neural Networks with Single Hidden Layers}

The complexity of pattern recognition problems can be quite different in practice. In Section 3.1, We first examine the different categories of classification problems  based on  the complexities of the patterns' separating boundaries. Then, we investigate the capacity of single hidden layer nets in Section 3.2.

\subsection{Separability of Pattern Sets}

Let $\mathcal{X}_1,\mathcal{X}_2$ be two disjoint pattern sets, that is, $\mathcal{X}_1 \cap \mathcal{X}_2 = \emptyset$. We introduce the following three categories of binary classification problems.   

\vskip 3mm
{\bf Linear-Separability of Two Categories}:  We say $\mathcal{X}_1$  is linearly-separable from $\mathcal{X}_2$ if there is a hyperplane in the vector space to separate $\mathcal{X}_1$ from $\mathcal{X}_2$. Note that, if $\mathcal{X}_1$  is linearly-separable from $\mathcal{X}_2$, then $\mathcal{X}_2$  is also linearly-separable from $\mathcal{X}_1$, linear-separability is mutual. It is also known that, $\mathcal{X}_1$ and $\mathcal{X}_2$ are linearly-separable if and only if  $\mathrm{CH}\{\mathcal{X}_1\} \cap \mathrm{CH}\{\mathcal{X}_2\}=\emptyset$.

\vskip 3mm  
{\bf Unidirectional Convex-Separability of Two Categories}: $\mathcal{X}_1$  is called convexly-separable from $\mathcal{X}_2$ if there is a convex region including all the points in $\mathcal{X}_1$ while excluding all the points in $\mathcal{X}_2$. $\mathcal{X}_1$ is convexly-separable from $\mathcal{X}_2$  if and only if  $\mathrm{CH}\{\mathcal{X}_1\} \cap \mathcal{X}_2=\emptyset$.

\vskip 3mm
{\bf Mutual Convex-Separability of Two Categories}: $\mathcal{X}_1$  and $\mathcal{X}_2$ are called mutually convexly-separable from each other if each of them is convexly-separable from the other. They are mutually convexly-separable if and only if $\mathrm{CH}\{\mathcal{X}_1\} \cap \mathcal{X}_2=\emptyset$ and $\mathrm{CH}\{\mathcal{X}_2\} \cap \mathcal{X}_1=\emptyset$. Note that mutual convex-separability is weaker than linear separability, that is, any two linearly-separable pattern sets are mutually convexly-separable but mutually convexly-separable pattern sets may not be linearly-separable.

\subsubsection{Separability of Multiple Pattern Sets}

For multiple category data sets $\mathcal{X}_1,\mathcal{X}_2, \cdots, \mathcal{X}_m $, an $m$-dimensional function is usually used to classify the patterns. We call $\mathbf{f}(\mbx)=[f_1(\mbx),f_2(\mbx),\cdots, f_m(\mbx)]$ an $m-$dimensional separator of $m$ disjoint pattern sets $\{\mathcal{X}_k,k\in [m]\}$ if, for each $k\in [m]$,  $f_k(\mbx)>0$ for all $\mbx\in \mathcal{X}_k$ and  $f_k(\mbx)<0$ for all $\mbx\in \bigcup_{j\not=k}\mathcal{X}_j$.   

\vskip 3mm
{\bf Pairwise Linear-Separability of Multiple Categories}:  We say $m$ pattern sets $\{\mathcal{X}_i\}_{i=1}^m$ are pairwise linearly-separable if every two pattern sets are linearly-separable, that is,
\begin{equation}
\mathrm{CH}\{\mathcal{X}_i\}\cap \mathrm{CH}\left\{\mathcal{X}_j\right\}=\emptyset, \forall\; i\not= j.
\end{equation} 

{
\vskip 3mm
{\bf Pairwise Convex-Separability of Multiple Categories}:  We say $m$ pattern sets $\{\mathcal{X}_i\}_{i=1}^m$ are pairwise convexly-separable if every two pattern sets are mutually convexly-separable, that is,
\begin{equation}
\begin{array}{rcl}
\mathrm{CH}\{\mathcal{X}_i\} &\cap& \mathcal{X}_j = \emptyset, \forall\; i\not= j\\
\mathrm{CH}\{\mathcal{X}_j\} &\cap& \mathcal{X}_i = \emptyset, \forall\; i\not= j.
\end{array}
\end{equation} 
}

\vskip 3mm
{\bf Linear-Separability of Multiple Categories}:  We say $m$ pattern sets $\{\mathcal{X}_i\}_{i=1}^m$ are linearly-separable if every pattern set $\mathcal{X}_i$ is linearly-separable from the union of all other pattern sets, that is,
\begin{equation}
\mathrm{CH}\{\mathcal{X}_i\}\cap \mathrm{CH}\left\{\cup_{j\not=i}\mathcal{X}_j\right\}=\emptyset.
\end{equation} 
According to this definition, there exists an $m$-dimensional linear classifier such that each pattern is positive in one axis and all of the other patterns are negative in this axis.  

For multiple category classification, linear separability is much stronger than pairwise linear separability.

\subsubsection{Separability of Multiple Pattern Sets with ReNN}

Given $m$ pattern sets, namely $\mathcal{X}_i, i\in [m]$, we call them separable by ReNNs, which may have additional sign constraints, if an ReNN (with corresponding constraints if any) exists such that 
\begin{equation}
y_i(\mbx)\left\{ \begin{array}{ll} >0,\;&\forall \; \mbx\in \mathcal{X}_i\\
<0, \; &\forall \mbx\in \bigcup_{j\not= i} \mathcal{X}_j
\end{array}\right.
\end{equation}
for all $i\in [m]$.

\subsection{Binary Classification Capacity of Single Hidden Layer Networks}

For binary classification, we only need one dimensional classifiers, and the sign-constrained ReNN defined in Eq. (\ref{SCReNN1}) can be described as
\begin{equation}
f(\mathbf{x})\triangleq \mathbf{a}^T\max(\mathbf{0},W^T\mathbf{x}+\mathbf{b})+c
\label{singleHiddenLayer}
\end{equation}
where the output layer weight matrix $A$ is reduced to a vector $\mathbf{a}\in\mathbb{R}^m$ and the bias vector $\mathbf{c}$ is reduced to be a scalar $c$.

Next, we establish the connections between sign constrained rectifier networks and convexly-separable pattern sets. For pattern sets $\mathcal{X}_{+}$ and $\mathcal{X}_{-}$ labelled positive and negative respectively,  a single-hidden-layer binary classifier $f(\mbx)$, as defined in Eq. (\ref{singleHiddenLayer}), is called a single hidden layer separator of $\mathcal{X}_{+}$ and $\mathcal{X}_{-}$ if it satisfies
\begin{equation}
\begin{array}{rcl}
f(\mbx)&>&0, \;\forall \;\mbx\in \mathcal{X}_{+}\\
f(\mbx)&< &0, \;\forall \;\mbx\in \mathcal{X}_{-}.\\
\end{array}
\label{sepCondition}
\end{equation}
If it further satisfies $\mathbf{a}\preceq 0, c\geq 0$, we call it a {\it sign-constrained single-hidden-layer separator} of $\mathcal{X}_{+}$ and $\mathcal{X}_{-}$.

 \begin{lem} {\it Let $\mathcal{X}_{+},\mathcal{X}_{-}$ be a pair of finite pattern sets in $\mathbb{R}^n$ and be labelled positive and negative respectively. Then $\mathcal{X}_{+}, \mathcal{X}_{-}$ can be separated by a sign-constrained single-hidden-layer classifier, as defined in Eq. (\ref{singleHiddenLayer}) and satisfying $c\geq 0$ and $\mathbf{a}\preceq 0$, if and only if the positive pattern set $\mathcal{X}_{+}$ is convexly-separable from the negative pattern set $\mathcal{X}_{-}$, i.e., $\mathcal{X}_{-}\cap \mathrm{CH}(\mathcal{X}_{+})=\emptyset$.  
 }
 \label{sepSHL}
 \end{lem}

{\bf Proof}: (Sufficiency). Suppose $\mathrm{CH}(\mathcal{X}_{+})\cap \mathcal{X}_{-}=\emptyset$. Let $n_{-}$ be the number of training patterns in $\mathcal{X}_{-}$ and $\mbx_i^{-}$ be the $i^{th}$ member of $\mathcal{X}_{-}$. Since $\mbx_i^{-}\not\in \mathrm{CH}(\mathcal{X}_{+})$  for any $i\in [n_{-}]$, there exists $\mbw_i, b_i, i\in [n_{-}]$ such that 
\begin{equation}
\begin{array}{rcl}
\mbw_i^T\mbx_i^{-}+ b_i &>& 0\\
\mbw_i^T\mbx+b_i &< & 0, \;\forall\; \mbx\in \mathcal{X}_{+}.     
\end{array}
\end{equation}
Note that the responses of positive patterns are negative, making $\mathcal{X}_{+}$ shrinking into a single point $\mathbf{0}$. 

Denote  
\begin{equation}
\begin{array}{rcl}
W&=&[\mbw_1,\mbw_2,\cdots,\mbw_{n_{-}}]\\
\mathbf{b}&=&[b_1,b_2,\cdots,b_{n_{-}}]^T\\
\mbz &=& \max(\mathbf{0},W^T\mbx +\mathbf{b}).
\end{array}
\label{Wbz}
\end{equation}
 Then we have 
\begin{equation}
\begin{array}{rcl}
\mathcal{Z}_{+} & \triangleq& \{\mbz = \max(\mathbf{0},W^T\mbx +\mathbf{b}) : \mbx\in \mathcal{X}_{+}\} \\
&=& \{\mathbf{0}\}\\
\mathcal{Z}_{-} & \triangleq& \{\mbz = \max(0,W^T\mbx +\mathbf{b}) : \mbx\in \mathcal{X}_{-}\} \\
&\subset& \{\mbz: \mathbf{1}^T\mbz > \gamma_{\min},\mbz\not=\mathbf{0},\; z_i\geq 0, \forall\ i\in [n_{-}]. \}\\
\end{array}
\end{equation}
where
\begin{equation}
\begin{array}{rcl}
\gamma_{\min} &\triangleq & \displaystyle\min_{\mbx\in \mathcal{X}_{-}} \mathbf{1}^T \max(\mathbf{0},W^T\mbx +\mathbf{b})\\
&>& 0. 
\end{array}
\end{equation}

For a single-hidden-layer binary classifier $f(\mbx)$, as described in Eq. (\ref{singleHiddenLayer}), if we choose $c = 1$ and $\mathbf{a}= -\frac{2}{\gamma_{\min}}\mathbf{1}\preceq 0$, then  
\[
f(\mbx)=-\frac{2}{\gamma_{\min}}\mathbf{1}^T\max(0,W^T\mbx+\mathbf{b})+1
\] 
satisfies
\begin{equation}
\begin{array}{rclcl}
f(\mbx) &\leq & -1&<& 0, \;\forall \; \mbx\in \mathcal{X}_{-},\\
f(\mbx) &= & 1&>&0, \;\forall \; \mbx\in \mathcal{X}_{+}\\
\end{array}
\end{equation}
which implies that $\mathcal{X}_{+}$ and $\mathcal{X}_{-}$ can be separated by a sign-constrained single-hidden-layer binary classifier.

(Necessity). Suppose that $\mathcal{X}_{+},\mathcal{X}_{-}$ can be separated by a sign-constrained single-hidden-layer binary classifier with $\mathbf{a}\preceq 0, c\geq  0$ such that $f(\mbx)$, as defined in Eq. (\ref{singleHiddenLayer}), satisfies Eq. (\ref{sepCondition}). Next, we will prove the convexity of the set $\{ \mbx : f(\mbx)> 0\}$ and show that  $f(\mbx)> 0$ holds for all $\mbx$ in the convex hull of $\mathcal{X}_{+}$.

Let $z_0,z_1$ be two arbitrary real numbers and let $z_{\lambda}=\lambda z_1+(1-\lambda)z_0$ be their linear combination. Since
\begin{equation}
\max(0,z_{\lambda})\leq \lambda \max(0,z_1)+(1-\lambda)\max(0,z_0),\; \forall\ \lambda\in [0,1], 
\end{equation} 
we have 
\begin{equation}
\begin{array}{l}
\mathbf{a}^T\max(0,\mbz_{\lambda})\geq \lambda \mathbf{a}^T\max(0,\mbz_0)+(1-\lambda)\mathbf{a}^T\max(0,\mbz_1), \\
\hspace{6cm} \forall\ \lambda\in [0,1]
\end{array} 
\end{equation} 
for any $\mathbf{a}\preceq 0$ and $\mbz_0,\mbz_1$ with the same dimensions, where $\mbz_{\lambda}\triangleq \lambda \mbz_1+(1-\lambda)\mbz_0$.

In particular, let $\mbz_{\lambda}=W^T\mbx_{\lambda}+\mathbf{b}$ with $\mbx_{\lambda}=\lambda \mbx_1+(1-\lambda)\mbx_0$. Then we have  
\begin{equation}
\begin{array}{rcl}
f(\mbx_{\lambda}) &=& \mathbf{a}^T\max(0,\mbz_{\lambda})+\beta\\
&\geq& \lambda \left[ \mathbf{a}^T\max(0,\mbz_0)+\beta \right]+\\
&&(1-\lambda)\left[ \mathbf{a}^T\max(0,\mbz_1)+\beta \right]\\
 &=& \lambda f(\mbx_0) + (1-\lambda) f(\mbx_1), \forall\ \lambda\in [0,1] \\
\end{array}
\end{equation}
and therefore 
\begin{equation}
f(\mbx_{\lambda})> 0,\;\forall \;\lambda\in [0,1] 
\end{equation}
if and only if
\begin{equation}
f(\mbx_{\lambda})> 0,\;\forall \; \lambda=0,1. 
\end{equation}
Hence $\{ \mbx : f(\mbx)> 0\}$
is a convex set, and thus
\begin{equation}
\begin{array}{rcl}
f(\mbx)> 0, \forall\; \mbx\in \mathrm{CH}(\mathcal{X}_{+})\\
\end{array}
\label{shrnCondition2}
\end{equation}
follows from $f(\mbx)> 0, \forall\; \mbx\in \mathcal{X}_{+}$. Note that $f(\mbx)<0$ for all $\mbx\in \mathcal{X}_{-}$ (from Eq. (\ref{sepCondition})). So $\mathcal{X}_{-}$ and $\mathrm{CH}(\mathcal{X}_{+})$ are separable and thus  $\mathrm{CH}(\mathcal{X}_{+})\cap \mathcal{X}_{-}=\emptyset $, which completes the proof.

\begin{flushright}
$\Box$
\end{flushright}

\subsection{Data Decomposition}

The following Lemma shows the capacity of sign-constrained single-hidden-layer classifiers in decomposing the negative pattern set into several subsets so that each subset is linearly separable from the positive pattern set.  

\begin{lem}
{\it 
Let $\mathcal{X}_{+}$ be a pattern set which is convexly-separable from  $\mathcal{X}_{-}$, and let $f(\mbx)$ , as defined in Eq.(\ref{singleHiddenLayer}) with $l$ hidden nodes and satisfying $\mathbf{a}\preceq 0, c\geq 0$, be one of their sign-constrained single-hidden-layer separators. Define 
\begin{equation}
f_{\mathcal{I}}(\mbx)\triangleq \left(\sum_{i\in \mathcal{I}} a_i (\mbw_i^T\mbx +b_i)\right)+c
\end{equation} 
and 
\begin{equation}
\mathcal{X}_{-}^{\mathcal{I}} \triangleq \left\{ \mathbf{x} : f_{\mathcal{I}}(\mbx)<0, \mbx\in \mathcal{X}_{-}\right\}. 
\label{piecesOfXplus}
\end{equation} 
for any subset $\mathcal{I}\subset [l]$. Then we have 
\begin{equation}
{X}_{-} = \bigcup_{\mathcal{I}\subset [l]}\mathcal{X}_{-}^{\mathcal{I}}
\label{decomOfXplus}
\end{equation} 
and 
\begin{equation}
\mathrm{CH}\left(\mathcal{X}_{-}^{\mathcal{I}}\right)\cap \mathrm{CH}(\mathcal{X}_{+})=\emptyset,
\label{linSep}
\end{equation}
i.e, $\mathcal{X}_{-}^{\mathcal{I}}$ and $\mathcal{X}_{+}$ are linearly-separable, and furthermore,  $f_{\mathcal{I}}(\mbx)$ is their linear separator satisfying
\begin{equation}
\begin{array}{rcl}
f_{\mathcal{I}}(\mbx)&<&0, \;\forall \;\mbx\in \mathcal{X}_{-}^{\mathcal{I}}\\
f_{\mathcal{I}}(\mbx)&> &0, \;\forall \;\mbx\in \mathcal{X}_{+}.\\
\end{array}
\label{sepCondition4}
\end{equation}
}
\label{DecomSHL}  
\end{lem}

Before proving this Lemma, we give an example to explain the subsets of $[l]$. When $l=2$, $[l]$ has three subsets: $\mathcal{I}_{1}=\{1\}, \mathcal{I}_{2}=\{2\}$ and $\mathcal{I}_{3}=\{1,2\}$.  Note that the decomposed subsets may have overlaps. The number of subsets is determined by the number of hidden nodes. For compact decompositions and meaningful discriminate feature analysis, small numbers of hidden nodes are preferable. The significance of this Lemma is in the discovery that a single-hidden-layer SCReNN can decompose the convexly-separable pattern sets into linearly-separable subsets so that the discriminate features of convexly-separable patterns can be analysed through the linear classifiers separating one pattern set (labelled positive) from the subsets of the other pattern set (labelled negative).  

\vskip 2mm

{\bf Proof}: From $\mathbf{a}\preceq 0$, it follows that
\begin{equation}
f_{\mathcal{I}}(\mbx)\geq f(\mbx), \; \forall\; \mathcal{I}\subset [l], \mbx\in \mathbb{R}^n
\end{equation} 
and consequently
\begin{equation}
\begin{array}{rcl}
f_{\mathcal{I}}(\mbx) &> & 0, \; \forall\; \mathcal{I}\subset [l], \mbx\in \mathcal{X}_{+}.\\
\end{array}
\label{temp}
\end{equation}
Then Eq. (\ref{sepCondition4}) follows directly from Eq. (\ref{temp}) and the definition of $\mathcal{X}_{-}^{\mathcal{I}}$ in Eq. (\ref{piecesOfXplus}). Note that $f_{\mathcal{I}}(\mbx)$ is a linear classifier satisfying Eq. (\ref{sepCondition4}), $f_{\mathcal{I}}(\mbx)$ is a linear separator of $\mathcal{X}_{-}^{\mathcal{I}}$ and $\mathcal{X}_{+}$, and Eq. (\ref{linSep}) holds consequently.  
 
To complete the proof, it remains to prove Eq. (\ref{decomOfXplus}). Let $\mbx\in \mathcal{X}_{-}$ be any pattern with negative label and let $\mathcal{I}\subset [l]$ be the index set so that $\mbw_i^T\mbx+b_i>0$ for all $i\in \mathcal{I}$ and $\mbw_i^T\mbx+b_i\leq 0$ for all $i\not\in \mathcal{I}$. Then $f_{\mathcal{I}}(\mbx) =f(\mbx) < 0$ and thus $\mbx\in \mathcal{X}_{-}^{\mathcal{I}}$. This proves that any element in $\mathcal{X}_{-}$ is in $\mathcal{X}_{-}^{\mathcal{I}}$ for some $\mathcal{I}\subset [l]$. Hence Eq. (\ref{decomOfXplus}) is true and the proof is completed.

\begin{flushright}
$\Box$
\end{flushright}

{

\subsection{Multiple Category Classification with Single Hidden Layers}

 This section considers multiple category classification problems. We will show that multiple (equal to or more than three) sets can be transformed to be linearly separable by a single hidden layer {\it if and only if} every pair of the classes are mutually convexly-separable.
 
 For classification of $m$ classes, an $m$-dimensional classifier, namely $\mathbf{f}(\mbx)=[f_1(\mbx),f_2(\mbx),\cdots,f_m(\mbx)]$, with sign-constrained ReNN can be described as 
 \begin{equation}
 f_i(\mbx)=\mathbf{a}_i^T\max(0,W^T\mbx+\mathbf{b})+c_i, \mathbf{a}_i\preceq 0, i=1,2,\cdots,m.
 \label{SHLm}
 \end{equation}
 
\begin{thm} {\it Multiple pattern sets, namely $\mathcal{X}_i (i=1,2,\cdots, m)$ can be separated by a sign-constrained single-hidden-layer classifier, as defined in Eq. (\ref{SHLm}), if and only if these pattern sets are pairwise mutually convex-separable, i.e., $\mathrm{CH}(\mathcal{X}_i)\cap \mathcal{X}_j=\emptyset $ for any $i\not=j$.
 }
 \label{sepSHLm}
 \end{thm}

 {\bf Proof}: From $\mathrm{CH}(\mathcal{X}_i)\cap \mathcal{X}_j=\emptyset,\;\forall\; j\not=i$, it follows that
\begin{equation}
\mathrm{CH}(\mathcal{X}_i)\cap \{\cup_{j\not= i} \mathcal{X}_j\}=\emptyset, \;\forall\; i\in [m]
\end{equation}
which implies that each pattern set is convexly-separable from the union of all the other patterns. 

Let $\mathcal{X}_{+}=\mathcal{X}_i$ and $\mathcal{X}_{-}=\bigcup_{j\not= i}\mathcal{X}_j$. By {\it Lemma} \ref{sepSHL}, there is a sign-constrained single-hidden-layer classifier, namely
\begin{equation}
f_i(\mbx)\triangleq \mathbf{a}_i^T\max\{\mathbf{0}, W_i^T\mbx+\mathbf{b}_i\}+c_i
\end{equation}
 with $\mathbf{a}_i\preceq 0$ and $c_i\geq 0$, such that 
 \begin{equation}
 \begin{array}{rcl}
 f_i(\mbx)>0,&& \forall \mbx\in \mathcal{X}_i\\
 f_i(\mbx)<0,&& \forall \mbx\in \bigcup_{j\not= i} \mathcal{X}_j.
 \end{array}
 \end{equation}

Let the multiple output SCReNN1 in Eq. (\ref{SCReNN1}) be defined with
\begin{equation}
\begin{array}{rcl}
A &=& \left[\begin{array}{ccc}
\mathbf{a}_1 && \\ &\ddots&\\ && \mathbf{a}_m
\end{array}\right]\\
W&=& \left[W_1, W_2, \cdots, W_m\right]\\
\mathbf{b}&=& \left[\mathbf{b}_1^T, \mathbf{b}_2^T, \cdots, \mathbf{b}_m^T\right]^T\\
\mathbf{c}&=& \left[c_1, c_2, \cdots, c_m\right]^T.\\
\end{array}
\end{equation}
Then the output at the $i^{th}$ node, namely $y_i$, equals to $f_i(x)$, and therefore 
\begin{equation}
\begin{array}{rcl}
y_i>0, && \forall\; \mbx\in\mathcal{X}_i\\
y_i<0, && \forall\; \mbx\in \bigcup_{j\not=i} \mathcal{X}_j\\
\end{array}
\end{equation}
which are true for all $i\in [m]$. This proves the sufficiency of mutual convex-separability for multiple category pattern sets to be separable by a single-hidden-layer SCReNN. Next, we prove its necessity. Suppose a single-hidden-layer SCReNN exists to separate $m$ category pattern sets $\{\mathcal{X}_i\}_{i=1}^m$ such that 
\begin{equation}
y_i(\mbx)\triangleq \mathbf{a}_i^T\max\{\mathbf{0}, W^T\mbx+\mathbf{b}\}+c_i
\end{equation}
satisfies: $y_i(\mbx)>0$ for all $\mbx\in \mathcal{X}_i$ and $y_i(\mbx)<0$ for all $\mbx\in \bigcup_{j\not=i} \mathcal{X}_j$. Note that $\mathbf{a}_i\preceq 0$, $y_i(\mbx)$ is a sign constrained single-hidden-layer separator of $\mathcal{X}_i$ from the set $\bigcup_{j\not=i}\mathcal{X}_j$. Then by Lemma \ref{sepSHL}, $\mathcal{X}_i$ is convexly-separable from the union of all the other pattern sets, and therefore $\mathcal{X}_i$ is convexly-separable from any other pattern set $\mathcal{X}_j (j\not=i)$. Note that, this is true for all $i\in [m]$ and therefore every pair of the pattern sets are mutually convexly-separable, which completes the proof for the necessity of pairwise mutual convex-separability. 
\begin{flushright}
$\Box$
\end{flushright}

Note that each dimension of the $m$ dimensional multiple category classifier is a sign-constrained ReNN to separate one pattern set from the others, the decomposition property of such sign-constrained multiple category classifiers can be derived directly from Lemma \ref{DecomSHL}. 

}

\section{The Capacity of Sign-Constrained Rectifier Networks with Two Hidden Layers}

In this section, we first investigate the universal classification power of sign-constrained two-hidden-layer binary classifiers and their capacity to decompose one pattern set into smaller subsets so that each subset is convexly separable from the other pattern set. We then extend this result to multiple category classification problems.

\subsection{Binary Classification with Two Hidden Layers}

A two-hidden-layer binary classifier, with $n$ dimensional input, $l$ bottom hidden nodes, $m$ top hidden nodes and a single output, can be described by
\begin{equation}
\begin{array}{rcl}
f\{\mathbf{g}(\mbx)\} &=& \mathbf{a}^T \max(\mathbf{0},W_2^T\mathbf{g}(\mbx) +\mathbf{b}_2)+c\\ 
\mathbf{g}(\mbx)&=&\max(\mathbf{0},W_1^T\mbx+\mathbf{b}_1)
\end{array}
\label{THLRN}
\end{equation}
where $c$ is a scalar number, $\mathbf{a}\in \mathbb{R}^{l_2}, \mathbf{b}\in \mathbb{R}^{l_2}, \mathbf{b}_1\in \mathbb{R}^{l_1}, W_2\in \mathbb{R}^{l_1\times l_2}$ and $W_1\in\mathbb{R}^{n\times l_1}$.

We say that a two-hidden-layer binary classifier $f\{\mathbf{g}(\mbx)\}$, as defined in Eq. (\ref{THLRN}), is a two-hidden-layer separator of $\mathcal{X}_{+}$ and $\mathcal{X}_{-}$ if it satisfies
\begin{equation}
\begin{array}{rcl}
f\{\mathbf{g}(\mbx)\}&>&0, \;\forall \;\mbx\in \mathcal{X}_{+}\\
f\{\mathbf{g}(\mbx)\}&< &0, \;\forall \;\mbx\in \mathcal{X}_{-}.\\
\end{array}
\label{sepCondition2}
\end{equation}
If it further satisfies $\mathbf{a}\preceq 0, \beta\geq 0$ and $W_2\preceq 0, \mathbf{b}_2\succeq 0$, we call it a {\it sign-constrained two-hidden-layer separator} of $\mathcal{X}_{+}$ and $\mathcal{X}_{-}$.

 \begin{lem} {\it For any two disjoint pattern sets, namely $\mathcal{X}_{+}$ and $\mathcal{X}_{-}$, in $\mathbb{R}^n$, there exists a sign-constrained two-hidden-layer binary classifier $f\{\mathbf{g}(\mbx)\}$, as defined in Eq. (\ref{THLRN}) and satisfying $\mathbf{a}\preceq 0, c\geq 0, W_2\preceq 0, \mathbf{b}_2\succeq 0$, such that  $f\{\mathbf{g}(\mbx)\}>0$ for all $\mbx\in\mathcal{X}_{+}$ and $f\{\mathbf{g}(\mbx)\}< 0$ for all $\mbx\in\mathcal{X}_{-}$.
 }
 \label{sepTHL}
 \end{lem}

{\bf Proof}: Let 
\begin{equation}
\mathcal{X}_{+}=\bigcup_{i=1}^{L_1}\mathcal{X}_{+}^{i}, \mathcal{X}_{-}=\bigcup_{j=1}^{L_2}\mathcal{X}_{-}^{j}
\end{equation}
be the disjoint convex hull decomposition \citep{an2015can} of $\mathcal{X}_{+}$ and $\mathcal{X}_{-}$. Then we have
\begin{equation}
\mathrm{CH}(\mathcal{X}_{+}^{i})\cap \mathrm{CH}(\mathcal{X}_{-}^{j}) \not= \emptyset, \;\forall\; i\in [L_1], j\in [L_2]
\end{equation}
 which implies that
\begin{equation}
\mathcal{X}_{+} \cap \mathrm{CH}(\mathcal{X}_{-}^{i})  \not= \emptyset, \;\forall\; i\in [L_2].
\end{equation}
 
Apply Lemma \ref{sepSHL} on $\mathcal{X}_{-}^{i}$ and $\mathcal{X}_{+}$ where the former is treated as positive pattern set. Then, for each $i$, there exists a sign-constrained single-hidden-layer separator between  $\mathcal{X}_{+}$ and $\mathcal{X}_{-}^{i}$. More precisely, there exist $\mbw_i\preceq 0, b_i\geq 0, W_1,\mathbf{b}_1$ such that
\begin{equation}
\begin{array}{rcl}
g_i(\mbx) &<& 0, \;\forall \mbx\in \mathcal{X}_{+}\\
g_i(\mbx) &> & 0, \;\forall \mbx\in \mathcal{X}_{-}^{i}\\
\end{array}
\end{equation}   
 where
 \begin{equation}
 g_i(\mbx)\triangleq \mbw_i^T\max(0,W_1^T\mbx+\mathbf{b}_1)+b_i.
 \end{equation}
 
 Let $W_2=[\mbw_1,\mbw_2,\cdots,\mbw_{L_2}]\preceq 0$, $\mathbf{b}_2=[b_1,b_2,\cdots, b_{L_2}]^T\succeq 0$ and consider the transformation 
\begin{equation}
\mbz=\mathbf{g}(\mbx)\triangleq \max(0,W_2^T\max(0,W_1^T\mbx+\mathbf{b}_1)+\mathbf{b}_2).
\end{equation}

 Denote 
\begin{equation}
\begin{array}{rcl}
\mathcal{Z}_{+} &\triangleq& \{\mbz: \mbz=\mathbf{g}(\mbx), \mbx\in \mathcal{X}_{+}\}\\
&=&\{\mathbf{0}\}\\
\mathcal{Z}_{-} &\triangleq& \{\mbz: \mbz=\mathbf{g}(\mbx), \mbx\in \mathcal{X}_{-}\}\\
&\subset& \{\mbz: \mathbf{1}^T\mbz > \gamma_{\min},\mbz\not=0, z_i\geq  0, \forall i\in [L_1] \}\\
\end{array} 
\end{equation} 
where
\begin{equation}
\begin{array}{rcl}
\gamma_{\min}&\triangleq & \displaystyle\min_{\mbx\in \mathcal{X}_{-}} \mathbf{1}^T \max(0,W_2^T\mathbf{g}(\mbx) -\mathbf{b}_2)\\
&>& 0. 
\end{array}
\end{equation}

Let $\mathbf{a}= -\frac{2}{\gamma_{\min}}\mathbf{1}\succeq 0,c=1$ and $f(\mbz)\triangleq \mathbf{a}^T\mbz+c$. Then
 $f(\mbz)\leq -1 $ for any $\mbz\in \mathcal{Z}_{-}$ and $f(\mbz)=1 $ for any 
 $\mbz\in \mathcal{Z}_{+}$. Hence
\begin{equation}
f\{\mathbf{g}(\mbx)\}\triangleq \mathbf{a}^T\max(0,W_2^T\mathbf{g}(\mbx)+\mathbf{b}_2)+c
\end{equation} 
 satisfies: $f\{\mathbf{g}(\mbx)\}>0$ for $\mbx\in \mathcal{X}_{+}$ and $f\{\mathbf{g}(\mbx)\}< 0$ for $\mbx\in \mathcal{X}_{-}$.  Note that $W_2\preceq 0, \mathbf{b}_2\succeq 0, \mathbf{a}\preceq 0, c>0$, $f\{\mathbf{g}(\mbx)\}$ is a sign-constrained two-hidden-layer binary classifier, and the proof is completed.
 
 \begin{flushright}
 $\Box$
\end{flushright}

\subsection{Data Decomposition with Two Hidden Layers}
 
Next, we investigate the applications of the two-hidden-layer sign constrained ReNN classifier  to decompose one pattern set (labelled positive) into several subsets so that each subset is convexly separable from the other pattern set.

 \begin{lem} {\it Let $\mathcal{X}_{+},\mathcal{X}_{-}$ be two disjoint pattern sets and let $f\{\mathbf{g}(\mbx)\}$, as defined in Eq. (\ref{THLRN}) and satisfying $\mathbf{a}\preceq 0, c\geq 0, W_2\preceq 0, \mathbf{b}_2\succeq 0$, be one of their sign-constrained two-hidden-layer binary separators with $l_2$ top hidden nodes and satisfying Eq. (\ref{sepCondition2}). Let $\mbw_i$ denote the $i^{th}$ column of $W_2$, $b_i$ denote the $i^{th}$ element of $\mathbf{b}_2$, and define
\begin{equation}
\begin{array}{rcl}
f_{\mathcal{I}}\{\mathbf{g}(\mbx)\}&\triangleq&\left(\displaystyle\sum_{i\in \mathcal{I}} a_i [\mbw_i^T\mathbf{g}(\mbx) +b_i]\right)+c \\
\mathcal{X}_{-}^{\mathcal{I}} &\triangleq & \left\{ \mathbf{x} : f_{\mathcal{I}}\{\mathbf{g}(\mbx)\}<0, \mbx\in \mathcal{X}_{-}\right\} 
\end{array}
\label{XIplus}
\end{equation} 
for any subset, namely $\mathcal{I}$, in $[l_2]$. Then we have 
\begin{equation}
{X}_{-} = \bigcup_{\mathcal{I}\subset [m]}\mathcal{X}_{-}^{\mathcal{I}}
\label{XplusDecom}
\end{equation} 
and 
\begin{equation}
\mathrm{CH}\left(\mathcal{X}_{-}^{\mathcal{I}}\right)\cap \mathcal{X}_{+}=\emptyset,
\label{convexSep}
\end{equation}
i.e, $\mathcal{X}_{-}^{\mathcal{I}}$ is convexly-separable from $\mathcal{X}_{+}$. Furthermore,  $f_{\mathcal{I}}\{\mathbf{g}(\mbx)\}$ is their single-hidden-layer  separator satisfying 
\begin{equation}
\begin{array}{rcl}
f_{\mathcal{I}}\{\mathbf{g}(\mbx)\}&>&0, \;\forall \;\mbx\in \mathcal{X}_{+}\\
f_{\mathcal{I}}\{\mathbf{g}(\mbx)\}&< &0, \;\forall \;\mbx\in \mathcal{X}_{-}^{\mathcal{I}}.\\
\end{array}
\label{sepCondition3}
\end{equation}
 }
 \label{Decom_THL}
 \end{lem}

{\bf Proof}: Note that the elements of $\mathbf{a}$ are zero or negative, i.e., $a_i\leq 0$, we have
\begin{equation}
f_{\mathcal{I}}\{\mathbf{g}(\mbx)\}\geq f\{\mathbf{g}(\mbx)\}, \; \forall \; \mbx\in \mathbb{R}^n
\end{equation}
and therefore
\begin{equation}
f_{\mathcal{I}}\{\mathbf{g}(\mbx)\} > 0, \; \forall \; \mbx\in \mathcal{X}_{+}
\end{equation}
which proves the first inequality of Eq. (\ref{sepCondition3}) while the second one follows from the definition of  $\mathcal{X}_{-}^{\mathcal{I}}$ in Eq. (\ref{XIplus}). Hence, $f_{\mathcal{I}}\{\mathbf{g}(\mbx)\}$ is a single-hidden-layer  separator of $\mathcal{X}_{+}$ and $\mathcal{X}_{-}^{\mathcal{I}}$. 
Note that
\begin{equation}
\begin{array}{rcl}
f_{\mathcal{I}}\{\mathbf{g}(\mbx)\}
&=&\mathbf{a}_{\mathcal{I}}^T\mathbf{g}(\mbx)+c_{\mathcal{I}}\\
&=& \mathbf{a}_{\mathcal{I}}^T\max(0,W_1^T\mbx+\mathbf{b}_1)+c_{\mathcal{I}}
\end{array}
\end{equation}
where
\begin{equation}
\begin{array}{rcl}
\mathbf{a}_{\mathcal{I}}&=&\displaystyle\sum_{i\in \mathcal{I}} a_i\mathbf{w}_i\succeq 0\\
c_{\mathcal{I}}&=&\displaystyle\sum_{i\in \mathcal{I}} a_ib_i+c. 
\end{array}
\end{equation} 
which imply that $(-f_{\mathcal{I}}\{\mathbf{g}(\mbx)\})$ is a {\it sign-constrained} single-hidden-layer  separator of $\mathcal{X}_{-}^{\mathcal{I}}$ from $\mathcal{X}_{+}$. Then by Lemma \ref{sepSHL}, we have Eq. (\ref{convexSep}). 
 
 Now it remains to prove Eq. (\ref{XplusDecom}). It suffices to prove that, for any $\mbx\in \mathcal{X}_{-}$, there exists $\mathcal{I}\subset [m]$ such that $\mbx\in \mathcal{X}_{-}^{\mathcal{I}}$.  Let $\mbx$ be a member in $\mathcal{X}_{-}$ and let $\mathcal{I}\subset [l_2], \mathcal{I}\not= \emptyset$ be the index set such that $\mbw_i^T\mathbf{g}(\mbx)+b_i>0$ for all $i\in \mathcal{I}$ and $\mbw_i^T\mathbf{g}(\mbx)+b_i\leq 0$ for all $i\not\in \mathcal{I}$. Then $f_{\mathcal{I}}\{\mathbf{g}(\mbx)\}=f\{\mathbf{g}(\mbx)\}  < 0$ and thus $\mbx$ is in $\mathcal{X}_{-}^{\mathcal{I}}$.
 \begin{flushright}
 $\Box$
 \end{flushright}

Lemma \ref{Decom_THL} states that the negative pattern set can be decomposed into several subsets by a two-hidden-layer SCReNN, namely 
\[
\mathcal{X}_{-} = \displaystyle\bigcup_{i=1}^t \mathcal{X}_{-}^i
\]
 so that each $\mathcal{X}_{-}^i$ is convexly-separable from $\mathcal{X}_{+}$. Then by labelling $\mathcal{X}_{-}^i$ as positive and $\mathcal{X}_{+}$ as negative, and from Lemma \ref{sepSHL}, $\mathcal{X}_{+}$ can be decomposed into a number, namely $t_i$, of subsets by a single-hidden-layer SCReNN, namely,
\[
\mathcal{X}_{+} = \displaystyle\bigcup_{j=1}^{t_i} \mathcal{X}_{+}^j,
\]
 so that $\mathcal{X}_{+}^i$ and $\mathcal{X}_{-}^j$ are linearly-separable. Hence, one can investigate the discriminant features of the two patterns by using the linear classifiers of these subsets of the patterns. With the decomposed subsets, one can investigate the pattern structures. The numbers of the subsets are determined by the numbers of hidden nodes in the top hidden layers of the two-hidden-layer SCReNNs and the numbers of the hidden nodes of the single-hidden-layer SCReNNs. To find compact pattern structures and meaningful discriminant features, a small number of hidden nodes are preferable.

 {

\subsection{Multiple Category Classification}

This section extends the results of the last section to multiple category classification problems. 
An $m$-dimensional classifier with two-hidden-layer $\mathrm{SCReNN}$, namely $\mathbf{f}(\mbx)=[f_1(\mbx),f_2(\mbx),\cdots, f_m(\mbx)]$,  can be described as
\begin{equation}
\begin{array}{rcl}
f_k(\mbx)&=&\mathbf{a}_i^T\max(0, W_2^T\max\{0,W_1^T\mbx+\mathbf{b}_1\}+\mathbf{b}_2)+\mathbf{c}\\
\mathbf{a}_k&\preceq& 0, W_2\preceq 0, k=1,2,\cdots,m\\
\end{array}
\label{THLm}
\end{equation}

\begin{thm}
 Let $\{\mathcal{X}_{k}, k=1,2,\cdots,m\}$ be $m$ disjoint pattern sets with a finite number of points, then there exists an $m$-dimensional classifier with two-hidden-layer $\mathrm{SCReNN}$, as defined in Eq. (\ref{THLm}), such that for each $k=1,2,\cdots,m$,$f_k(\mbx)$ is positive for any $\mbx\in \mathcal{X}_k$, and negative for any $\mbx$ from other pattern sets. 
\end{thm}

{\bf Proof}: Denote $\hat{\mathcal{X}}_k\triangleq \bigcup_{l=1,l\not=k}^m\mathcal{X}_l$. So $\mathcal{X}_{k}$ and  $\hat{\mathcal{X}}_k$ are disjoint. From Lemma \ref{sepTHL}, there exists two hidden layer SCReNNs 
 \begin{equation}
\begin{array}{rcl}
f_k\{G_k(\mbx)\} &=& \bar{\mathbf{a}}_k^T \max(\mathbf{0},U_k^TG_k(\mbx) +\mathbf{b}_{1,k})+c_k\\ 
G_k(\mbx)&=&\max(\mathbf{0},V_k^T\mbx+\mathbf{b}_{2,k})\\
\bar{\mathbf{a}}\preceq 0,&&U_k\preceq 0\\
\end{array}
\end{equation}
such that
\begin{equation}
\begin{array}{rcl}
f_k\{G_k(\mbx)\}&>&0, \;\forall \;\mbx\in \mathcal{X}_k\\
f_k\{G_k(\mbx)\}&< &0, \;\forall \;\mbx\in \hat{\mathcal{X}}_k.\\
\end{array}
\end{equation}
Denote 
\begin{equation}
\begin{array}{rcl}
W_1&\triangleq&[V_1,V_2,\cdots,V_m]\\
\mathbf{b}_1&\triangleq&[\mathbf{b}_{1,1}^T,\mathbf{b}_{1,2}^T,\cdots,\mathbf{b}_{1,m}^T]^T\\
\mathbf{b}_2&\triangleq&[\mathbf{b}_{2,1}^T,\mathbf{b}_{2,2}^T,\cdots,\mathbf{b}_{2,m}^T]^T\\
\mathbf{c}&\triangleq&[c_1,c_2,\cdots,c_m]^2\\
\end{array}
\label{W1stLayer}
\end{equation}
and
\begin{equation}
\begin{array}{rcl}
W_2&=& \left[\begin{array}{cccc} U_1&0&\cdots &0\\ 0&U_2&\cdots&0\\ \vdots &\vdots&\ddots&\vdots\\
0&0&\cdots&U_m\end{array}\right]\\
A&=&\left[\begin{array}{cccc} \bar{\mathbf{a}}_1&0&\cdots&0\\ 0&\bar{\mathbf{a}}_2&\cdots&0\\
\vdots&\vdots&\ddots&\vdots\\
0&0&\cdots&\bar{\mathbf{a}}_m
\end{array}\right].
\end{array}
\label{W2ndLayer}
\end{equation}
Now let the weights and bias of the two-hidden-layer ReNN, defined in Eq. (\ref{THLm}),  be chosen as in Eq. (\ref{W1stLayer}) and Eq. (\ref{W2ndLayer}), then the output $\mby$ of the network satisfies $y_k=f_k\{G_k(\mbx)\}$. Note that $A\preceq 0, W_2\preceq 0$. The defined ReNN is a two-hidden-layer sign-constrained ReNN and this complete the proof of the universal classification power of two-hidden-layer SCReNN.
\begin{flushright}
$\Box$
\end{flushright}

Next we address the decomposition capacity of two-hidden-layer rectifier neural networks. Suppose a two-hidden-layer SCReNN (as defined in Eq. (\ref{THLm})) is a separator of $m$ pattern sets $\mathcal{X}_i,i\in [m]$ such that the $i^{th}$ output $y_i$ satisfies $y_i(\mbx)>0$ for any $\mbx\in\mathcal{X}_i$ and $y_i(\mbx)<0$ for any $\mbx$ in the other pattern sets. Hence 
\begin{equation}
y_i(\mbx)=\mathbf{a}_i^T\max(0, W_2^T\max\{0,W_1^T\mbx+\mathbf{b}_1\}+\mathbf{b}_2)+\mathbf{c} 
\end{equation}
is a binary two-hidden-layer sign constrained ReNN separator for  
$\mathcal{X}_i$ and the set $\bigcup_{j\not=i}\mathcal{X}_j$. By Lemma \ref{Decom_THL}, one can decompose the union $\bigcup_{j\not=i}\mathcal{X}_j$ into a number of subsets that are convexly-separable from $\mathcal{X}_i$. Then, for each such subset, namely $\hat{\mathcal{X}}$, of the union, one can use a single-hidden-layer SCReNN to separate $\hat{\mathcal{X}}$ and $\mathcal{X}_i$, and decomposed $\mathcal{X}_i$ into several subsets which are linearly separable from $\hat{\mathcal{X}}$. One can use the linear separators of these decomposed subsets to analyse the data and the key discriminant factors. 
}


\section{Training of Sign Constrained Rectifier Networks}

In this section, we first introduce the well known MM algorithm for non-convex optimization problems, and then show how the convexity/concavity properties of SCRNs can be used to design convex surrogate functions in order to apply the MM algorithms to learn the parameters of the proposed SCRNs. 

\subsection{The MM Algorithm}

The MM algorithm is an iterative algorithm for minimization of non-convex objective functions, with each of its iterations consisting of two steps: the majorization step which finds a surrogate function that upperbounds the objective function, and the minimization step which minimizes the surrogate function. Suppose we have the following optimization problem
\begin{equation}
\min_{\mbx\in\Omega} f(\mbx)
\end{equation}
where $\Omega\subset \mathbb{R}^n$ is a closed convex set. The general idea of the MM procedure is to construct a majorization function $g(\mbx,\mbz)$ such that
\begin{equation}
\left\{\begin{array}{rcl}
f(\mbx)&\leq& g(\mbx,\mbz),\;\forall\; \mbx, \mbz\in\Omega\\
f(\mbx)&=& g(\mbx,\mbx), \;\forall\; \mbx\in \Omega
\end{array}\right.
\label{surBound}
\end{equation}
and update $\mbx$ at iteration $l$ by
\begin{equation}
\mbx^{(l+1)} = \mathrm{arg}\min_{\mbx\in\Omega} g(\mbx,\mbx^{(l)}).
\label{updat}
\end{equation}
 It is easy to show that the above
iterative scheme decreases the objective function  monotonically in each iteration, i.e.,
\begin{equation}
f(\mbx^{(l+1)}) \leq g(\mbx^{(l+1)}, \mbx^{(l)}) \leq g(\mbx^{(l)}, \mbx^{(l)}) = f(\mbx^{(l)}),
\end{equation}
where the first inequality and the last equality follow from (\ref{surBound}) while the sandwiched inequality
follows from (\ref{updat}). Hence the objective function decreases until it converges to a stationary point. Moreover, many local minima of the objective function can be avoided if they have larger values than the minimum of one of the convex functions in the iterations of the MM algorithm.  Hence, the MM algorithm usually finds a good solution even though it cannot guarantee to find the global minima.

Next, we use the MM algorithm to address the training of single hidden layer SCRNs. 

\subsection{Training of Single Hidden Layer SCRNs}

Consider the following single hidden layer SCRN 
\begin{equation}
\begin{array}{rcl}
f(\mbx;\mathcal{W})&=&b_0-\mathbf{1}^T\max\{0,W^T\mbx+\mbb\}\\
&=&b_0-\sum_{k=1}^m \max(0,\mbw_k^T\mbx+b_k)
\end{array} 
\end{equation}
for binary classification, where $\mathcal{W}\triangleq \{b_0,\mbw_k,b_k,k\in [m]\}$ is the set of weights and biases.  By Lemma 1, $f(\mbx;\mathcal{W})$ is a concave function of $\mathcal{W}$ when $\mbx$ is fixed.  Therefore, the hinge loss of the positive patterns 
\begin{equation}
\begin{array}{rcl}
J_{+}(\mathcal{W})&\triangleq& \dsum_{\mbx\in\mathcal{X}_{+}} \max\{0,1-f(\mbx;\mathcal{W})\}\\
\end{array}
\end{equation}
is a convex function of $\mathcal{W}$. Although the hinge loss of the negative patterns, namely
\begin{equation}
\begin{array}{rcl}
J_{-}(\mathcal{W})&\triangleq& \dsum_{\mbx\in\mathcal{X}_{-}} \max\{0,1+f(\mbx;\mathcal{W})\}\\
\end{array}
\end{equation}
is not convex, it is bounded by the following convex function
\begin{equation}
\begin{array}{rcl}
\widehat{J}_{-}(\mathcal{W};\mathcal{W}_0)&\triangleq& \dsum_{\mbx\in\mathcal{X}_{-}} \max\{0,1+\widehat{f}(\mbx;\mathcal{W},\mathcal{W}_0)\}\\
\end{array}
\end{equation}
where $\mathcal{W}_0=\{b_{0,0},\mbw_{k,0}, b_{k,0}, k\in [m]\}$ is a fixed set of parameters, and 
\begin{equation}
\begin{array}{rcl}
\widehat{f}(\mbx;\mathcal{W},\mathcal{W}_0)&\triangleq&b_0-\dsum_{k\in \mathcal{K}(\mbx, \mathcal{W}_0)} \mbw_k^T\mbx+b_k\\
&\geq& b_0-\dsum_{k\in [m]}\max\{0, \mbw_k^T\mbx+b_k\} \\ &=& f(\mbx;\mathcal{W})\\
\mathcal{K}(\mbx, \mathcal{W}_0)&\triangleq& \left\{k: 1\leq k\leq m,\mbw_{k,0}^T\mbx+b_{k,0}>0\right\}. 
\end{array}
\end{equation}
Therefore the total hinge loss satisfies the following inequality
\begin{equation}
\begin{array}{rcl}
J(\mathcal{W})&=& J_{+}(\mathcal{W})+J_{-}(\mathcal{W})\\
&\leq&J_{+}(\mathcal{W})+\widehat{J}_{-}(\mathcal{W};\mathcal{W}_0).
\end{array}
\end{equation}
That is, $J(\mathcal{W})$ is bounded by a convex function of $\mathcal{W}$, i.e., $J_{+}(\mathcal{W})+\widehat{J}_{-}(\mathcal{W};\mathcal{W}_0)$.

Hence, the minimization problem of the hinge loss $J(\mathcal{W})$, with  some convex regularization term $R(\mathcal{W})$, can be solved with the efficient MM algorithm as below
\begin{equation}
\begin{array}{rl}
\mathcal{W}^{(l+1)}=\mathrm{arg}\dmin_{\mathcal{W}}& R(\mathcal{W})+J_{+}(\mathcal{W})+\widehat{J}_{-}(\mathcal{W};\mathcal{W}^{(l)}).\end{array}
\label{SolverSHL} 
\end{equation}
At each iteration, one needs to solve a convex minimization problem and the cost function decreases until convergence, i.e.,
\begin{equation}
\begin{array}{l}
R(\mathcal{W}^{(l+1)})+J_{+}(\mathcal{W}^{(l+1)})+J_{-}(\mathcal{W}^{(l+1)})\\ \hspace{2cm} \leq R(\mathcal{W}^{(l)})+J_{+}(\mathcal{W}^{(l)})+J_{-}(\mathcal{W}^{(l)}).
\end{array}
\end{equation}

As many local minima of the non-convex function $J(\mathcal{W})$ can be avoided in the MM algorithm, the risk of local minima problem can be greatly reduced. Next, we consider the training of SCRNs with two hidden layers.

\subsection{Training of Two-Hidden-Layer SCRNs}

Let
\begin{equation}
\begin{array}{rcl}
f(\mbx;\mathcal{W})&=&b_0-\mathbf{1}^T\max(0,\mbz_2)\\
\mbz_2&=& W_2^T\max(0,\mbz_1)+\mbb_2\\
\mbz_1&=&W_1^T\mbx+\mbb_1\\
W_2&\preceq& 0
\end{array} 
\end{equation}
be a two-hidden-layer SCRN. We learn its weights and biases by minimizing the following cost function
\begin{equation}
\begin{array}{rcl}
J(\mathcal{W})&\triangleq &R(\mathcal{W})+J_{+}(\mathcal{W})+J_{-}(\mathcal{W})\\
J_{+}(\mathcal{W})&\triangleq& \dsum_{\mbx\in\mathcal{X}_{+}} \max\left\{0,1-f(\mbx;\mathcal{W})\right\}\\
J_{-}(\mathcal{W})&\triangleq& \dsum_{\mbx\in\mathcal{X}_{-}} \max\left\{0,1+f(\mbx;\mathcal{W})\right\}
\end{array}
\end{equation}
where $\mathcal{W}\triangleq \{b_0,W_1,W_2,\mbb_1,\mbb_2\}$ is the set of the parameters in the network, $R(\mathcal{W})$ is a convex regularisation term, $\mathcal{X}_{+}$ and $\mathcal{X}_{-}$ are the pattern sets with labels 1 and -1 respectively. 

When the first layer weights $W_1$ and biases $\mbb_1$ are fixed, the learning of the other parameters in $\mathcal{W}$ is essentially a training problem of a single hidden layer SCRN and thus can be optimized by using the algorithm presented in Section 5.2. Next, we consider the optimization of $(W_1,\mbb_1)$ when the other parameters are fixed. The following Lemma provides the foundation for the algorithm to be presented. 

\begin{lem}  
 Let $\mba_1\in\{0,1\}^{l_1},\mba_2\in\{0,1\}^{l_2}$ be two arbitrary activation patterns for the first layer nodes  and for the second layer nodes respectively. Denote 
\begin{equation}
\begin{array}{rcl}
f_{1}(\mbx;\mathcal{W},\mba_1)&=&b_0-\mathbf{1}^T\max(0,\hat{\mbz}_2)\\
\hat{\mbz}_2&=& W_2^T\mathrm{diag}\{\mba_1\}\mbz_1+\mbb_2\\
\mbz_1&=&W_1^T\mbx+\mbb_1\\
\end{array} 
\end{equation}
and 
\begin{equation}
\begin{array}{rcl}
f_{2}(\mbx;\mathcal{W},\mba_2)&=&b_0-\mba_2^T\mbz_2\\
\mbz_2&=& W_2^T\max(0,\mbz_1)+\mbb_2\\
\mbz_1&=&W_1^T\mbx+\mbb_1.\\
\end{array} 
\end{equation}
Then we have
\begin{enumerate}
\item[(i)] $f_{1}(\mbx;\mathcal{W},\mba_1)$ is a concave function of $(W_1,\mbb_1)$ when the other parameters in $\mathcal{W}$ are fixed, and furthermore
\begin{equation}
f_{1}(\mbx;\mathcal{W},\mba_1)\leq f(\mbx;\mathcal{W}).
\label{f1Ineq} 
\end{equation}
\item[(ii)] $f_{2}(\mbx;\mathcal{W},\mba_2)$ is a convex function of $(W_1,\mbb_1)$ when the other parameters in $\mathcal{W}$ are fixed, and furthermore 
\begin{equation}
f_{2}(\mbx;\mathcal{W},\mba_2)\geq f(\mbx;\mathcal{W}). 
\label{f2Ineq}
\end{equation}
\end{enumerate}
\label{THLoptLemma}
\end{lem}

{\bf Proof}: Note that $\hat{\mbz}_2$ is a linear function of $(W_1,\mbb_1)$ and $-\mathbf{1}\preceq 0$. From Lemma 1, it follows that $f_1(\mbx;\mathcal{W},\mba_1)$ is a concave function of $(W_1,\mbb_1)$. Furthermore, since $W_2\preceq 0$ and  $\mathrm{diag}\{\mba_1\}\mbz_1\leq \max(0,\mbz_1)$, we have $\hat{\mbz}_2\succeq \mbz_2$ and therefore Eq. (\ref{f1Ineq}) holds. This proves the first statement $(i)$.

For the proof of statement $(ii)$, Eq. (\ref{f2Ineq}) is true due to the fact that $\mba_2^T\mbz_2\leq \mathbf{1}^T\max(0,\mbz_2)$. To prove the convexity of  $f_{2}(\mbx;\mathcal{W},\mba_2)$ as a function of $(W_1,\mbb_1)$, let $\hat{b}_0=b_0-\mba_2^T\mbb_2$ and $\hat{\mba}_2^T=-\mba_2^TW_2\succeq 0$. Then $f_{2}(\mbx;\mathcal{W},\mba_2)=\hat{b}_0+\hat{\mba}_2\max(0,W_1^T\mbx+\mbb_1)$. Therefore, by Lemma 1,  $f_{2}(\mbx;\mathcal{W},\mba_2)$ is a convex function of $(W_1,\mbb_1)$ when $\mbx$ and the other parameters in $\mathcal{W}$ are fixed. This proves the statement $(ii)$ and completes the proof.
\begin{flushright}
$\Box$
\end{flushright}    

Based on Lemma \ref{THLoptLemma}, $(W_1,\mbb_1)$ can be optimized iteratively using the MM algorithm as follows. Let $\mathcal{W}^{(l)}$ be the parameter set at step $l$. $\mathcal{W}^{(0)}$ can be any arbitrary initialization. Let $\mathbf{a}_{1,l}(\mbx)$ and $\mathbf{a}_{2,l}(\mbx)$ be the activation patterns of $\mbz_1(\mbx)$ and $\mbz_2(\mbx)$ respectively at step  $l$, and denote
 
 \begin{equation}
\begin{array}{rcl}
\widehat{J}_{+}(\mathcal{W},\mathcal{W}^{(l)})&\triangleq& \dsum_{\mbx\in\mathcal{X}_{+}} \max\{0,1-f_1(\mbx;\mathcal{W},\mba_{1,l}(\mbx))\}\\
\widehat{J}_{-}(\mathcal{W},,\mathcal{W}^{(l)})&\triangleq& \dsum_{\mbx\in\mathcal{X}_{-}} \max\{0,1+f_2(\mbx;\mathcal{W},\mba_{2,l}(\mbx))\}.\\
\end{array}.
\end{equation}

Then, from Lemma 10, we have
\begin{equation}
\begin{array}{rcl}
J_{+}(\mathcal{W})&\leq& \widehat{J}_{+}(\mathcal{W},\mathcal{W}^{(l)})\\
J_{-}(\mathcal{W})&\leq& \widehat{J}_{-}(\mathcal{W},\mathcal{W}^{(l)}).\\
\end{array}
\label{Inequalities}
\end{equation}
Furthermore, $\widehat{J}_{+}(\mathcal{W},\mathcal{W}^{(l)})$ and $\widehat{J}_{-}(\mathcal{W},\mathcal{W}^{(l)})$ are convex functions of $(W_1,\mbb_1)$ when the other parameters are fixed. Hence $\mathcal{W}$ can be updated as
\begin{equation}
\begin{array}{rl}
\mathcal{W}^{(l+1)}=\mathrm{arg}\dmin_{W_1,\mbb_1} &  R(\mathcal{W})+\widehat{J_{+}}(\mathcal{W};\mathcal{W}^{(l)})+\widehat{J_{-}}(\mathcal{W};\mathcal{W}^{(l)}).\\
\end{array}
\label{maxOutIteration}
\end{equation}

From Eq. (\ref{Inequalities}), the cost function is strictly decreasing until convergence, that is
\begin{equation}
\begin{array}{l}
R(\mathcal{W}^{(l+1)})+J_{+}(\mathcal{W}^{(l+1)})+J_{-}(\mathcal{W}^{(l+1)})\\ \hspace{2cm}\leq R(\mathcal{W}^{(l)})+J_{+}(\mathcal{W}^{(l)})+J_{-}(\mathcal{W}^{(l)}).
\end{array} 
\end{equation} 

The whole set of parameters $\mathcal{W}$ can be learnt by optimizing $(b_0,W_2,\mbb_2)$ and $(W_1,\mbb_1)$ alternatively.

\section{Concluding Remarks}

We have shown that, with sign constraints on the weights of the output and the second hidden layers, two-hidden-layer SCRNs are still universal classifiers capable of  decomposing each class of patterns into several subsets so that each subset is convexly separable from the other pattern set. In addition, single-hidden-layer SCRNs are capable of separating any two (or more) convexly separable pattern sets as well as decomposing one of them into several subsets so that each subset is linearly separable from the other pattern set.  The proposed SCRN not only enables pattern and feature analysis for model interpretability and knowledge discovery but also enables efficient training with the well known MM algorithms to reduce the risks of local minima.

\section*{References}

\bibliography{icml2017bib}
\bibliographystyle{elsarticle-num} 








\end{document}